\pdfoutput=1

\documentclass[11pt]{article}

\usepackage[]{acl}

\usepackage{times}
\usepackage{latexsym}
\usepackage{booktabs}
\usepackage{xcolor,colortbl}
\usepackage{graphicx}
\usepackage{multirow}
\usepackage{enumitem}
\newcommand\Mark[1]{\textsuperscript#1}
\usepackage[T1]{fontenc}

\usepackage[utf8]{inputenc}

\usepackage{microtype}

\title{Benchmarking for Public Health Surveillance tasks on Social Media with a Domain-Specific Pretrained Language Model}

\author{Usman Naseem\Mark{1}, Byoung Chan Lee\Mark{2}, Matloob Khushi\Mark{1}, Jinman Kim\Mark{1}, Adam G. Dunn\Mark{3} \\
        \Mark{1}School of Computer Science, University of Sydney, Australia \\ \Mark{2}School of Medicine, University of Sydney, Australia \\
        \Mark{3}School of Medical Sciences, University of Sydney, Australia\\
        \texttt{blee9781@uni.sydney.edu.au}\\
        \texttt{\{usman.naseem,matloob.khushi,jinman.kim,adam.dunn\}@sydney.edu.au}}

\begin{document}
\maketitle
\begin{abstract}

A user-generated text on social media enables health workers to keep track of information, identify possible outbreaks, forecast disease trends, monitor emergency cases, and ascertain disease awareness and response to official health correspondence. This exchange of health information on social media has been regarded as an attempt to enhance public health surveillance (PHS). Despite its potential, the technology is still in its early stages and is not ready for widespread application. Advancements in pretrained language models (PLMs) have facilitated the development of several domain-specific PLMs and a variety of downstream applications. However, there are no PLMs for social media tasks involving PHS. We present and release PHS-BERT, a transformer-based PLM, to identify tasks related to public health surveillance on social media. We compared and benchmarked the performance of PHS-BERT on 25  datasets from different social medial platforms related to 7 different PHS tasks. Compared with existing PLMs that are mainly evaluated on limited tasks, PHS-BERT achieved state-of-the-art performance on all 25 tested datasets, showing that our PLM is robust and generalizable in the common PHS tasks. By making PHS-BERT available\footnote{https://huggingface.co/publichealthsurveillance/PHS-BERT}, we aim to facilitate the community to reduce the computational cost and introduce new baselines for future works across various PHS-related tasks.

\end{abstract}

\section{Introduction}

Public health surveillance (PHS) is defined by the World Health Organization\footnote{https://www.euro.who.int/en/health-topics/Health-systems/public-health-services} as the ongoing, systematic collection, assessment, and understanding of health-related required information for the planning, implementation, and assessment of healthcare~\cite{aiello2020social}. PHS aims to design and assist interventions; it acts as a primary warning system in health emergencies (epidemics, i.e., acute events), it reports and records public health interventions (i.e., monitoring health), and it observes and explains the epidemiology of health issues, allowing for the prioritization of necessary details for health policy formulation (i.e., targeting chronic events). Traditional PHS systems are often limited by the time required to collect data, restricting the quick or even instantaneous identification of outbreaks~\cite{hope2006syndromic}. 

Social media is growingly being used for public health purposes and can disseminate disease risks and interventions and promote wellness and healthcare policy. Social media data provides an abundant source of timely data that can be used for various public health applications, including surveillance, sentiment analysis, health communication,  and analyzing the history of a disease, injury, or promote health. Systematic reviews of studies that examine personal health experiences shared online reveal the breadth of application domains, which include infectious diseases and outbreaks~\cite{charles2015using}, illicit drug use~\cite{kazemi2017systematic}, and pharmacovigilance support~\cite{golder2015systematic}. These applied health studies are motivated by their potential in supporting PHS, augmenting adverse event reporting, and as the basis of public health interventions~\cite{dunn2018social}.

The use of deep learning in natural language processing (NLP) has advanced the development of pretrained language models (PLMs) that can be used for a wide range of tasks in PHS. However, directly applying the state-of-the-art (SOTA) PLMs such as Bidirectional Encoder Representations from Transformers (BERT)~\cite{devlin-etal-2019-bert}, and its variants~\cite{robertaliu2019roberta,lan2019albert,sanh2019distilbert,naseem2021comprehensive} that are trained on general domain corpus (e.g., Bookcorpus, Wikipedia,  etc.) may yield poor performances on domain-specific tasks. To address this limitation, several domain-specific PLMs have been presented. Some of the well-known in the biomedical field include the following: biomedical BERT (BioBERT)~\cite{lee2019biobert} and biomedical A Lite BERT (BioALBERT)~\cite{naseem2020bioalbert,naseem2021benchmarking}. Recently, other domain-specific LMs such as BERTweet~\cite{nguyen2020bertweet} for 3 downstream tasks, i.e.,  part-of-speech 
tagging, named-entity-recognition, and text classification and COVID Twitter BERT (CT-BERT)~\cite{muller2020covid} for 5 text classification tasks have been trained on datasets from Twitter. 

Despite the number of PLMs that have been released, none have been produced specifically for PHS from online text. Furthermore, all these LMs were evaluated with the selected dataset, and therefore their generalizability is unproven. To benchmark and fill the gap, we present PHS-BERT,  a new domain-specific contextual PLM trained and fine-tuned to achieve benchmark performance on various PHS tasks on social media. PHS-BERT is trained on a  health-related corpus collected from user-generated content. Our work is the first large-scale study to train, release and test a domain-specific PLM for PHS tasks on social media. We demonstrated that PHS-BERT outperforms other SOTA PLMs on 25 datasets from different social media platforms related to 7 different PHS tasks, showing that PHS is robust and generalizable.




\section{Related Work} \label{rw}

\subsection{Pretrained Language Models}

Transformer-based PLMs such as BERT~\cite{devlin-etal-2019-bert} and its variants~\cite{robertaliu2019roberta,lan2019albert} have altered the landscape of research in NLP domain. These PLMs are trained on a huge corpus but may not provide a good representation of specific domains~\cite{muller2020covid}. To improve the performance in domain-specific tasks, various domain-specific PLMs have been presented. Some of the famous in the biomedical domain are   BioBERT~\cite{lee2019biobert} and BioALBERT~\cite{naseem2020bioalbert}. Recently, for tasks on social media-specific, other PLMs such as BERTweet~\cite{nguyen2020bertweet}, COVID Twitter BERT (CT-BERT)~\cite{muller2020covid} have been trained on datasets from Twitter. For various downstream tasks, these domain-specific PLMs were demonstrated to be effective alternatives for PLMs trained on a general corpus for a variety of downstream tasks~\cite{muller2020covid}. The assumption is that the LMs trained on the user-generated text on Twitter can handle the short and unstructured text in tweets. Despite this progress, their generalizability is unproven, and there is no PLM for public health surveillance using social media.

\subsection{NLP for Public Health Surveillance}

The use of social media in conjunction with advances in NLP for PHS tasks is a growing area of study~\cite{paul2017social}. NLP can assist researchers in the surveillance of mental disorders, such as identifying depression diagnosis, assessing suicide risk and stress identification, vaccine hesitancy and refusal, identifying common health-related misconceptions, sentiment analysis, and the health-related behaviors they support~\cite{naseemdepression,naseemHMC}.

\begin{figure*}[!htpb]
\centering
\includegraphics[width=1\linewidth]{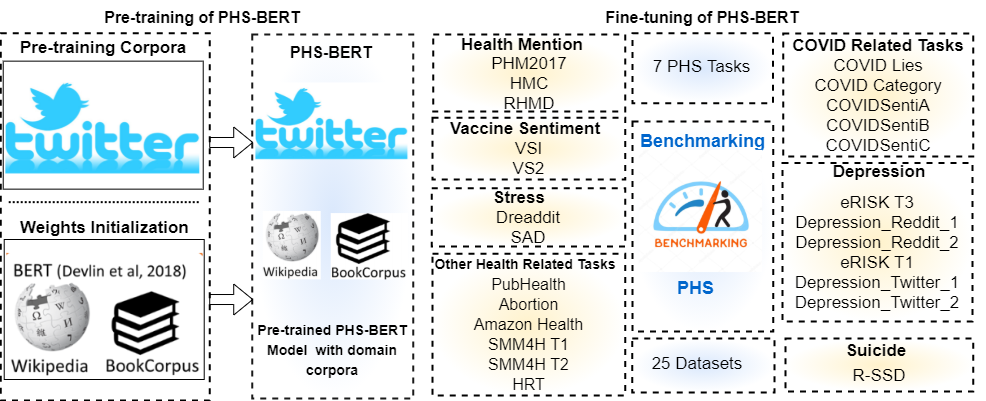}
\vspace{-0.70cm}
\caption{An overview of pretraining, fine-tuning, and the various tasks and datasets used in PHS benchmarking}
\label{ben}
\vspace{-0.4cm}
\end{figure*} 

\citet{rao2020knowledge} presented a hierarchical method that used BERT with attention-based BiGRU and achieved competitive performance for depression detection. For vaccine-related sentiment classification, \citet{zhang2020sentiment} classified tweet-level HPV vaccine sentiment using three transfer learning techniques (ELMo, GPT, and BERT) and found that a finely tuned BERT produced the best results. \citet{biddle2020leveraging} presented a method (BiLSTM-Senti)  that leveraged contextual word embeddings (BERT) with word-level sentiment to improve performance. \citet{naseem2021classifying} presented a model that uses domain-specific LM and captures commonsense knowledge into a context-aware bidirectional gated recurrent network. \citet{sawhney2021towards} presented an ordinal hierarchical attention model for Suicide Risk Assessment where text embeddings obtained by Longformer were fed to BiLSTM with attention and ordinal loss as an objective function. However, there is no PLM trained on health-related text collected from social media that directly benefit the applications related to PHS.
\section{Method} \label{aaamodel}

PHS-BERT has the same architecture as BERT. Fig.~\ref{ben} illustrates an overview of pretraining, fine-tuning, and datasets used in this study.   We describe BERT and then the pretraining and fine-tuning process employed in PHS-BERT.



\subsection{BERT}
PHS-BERT has the same architecture as BERT. 
BERT was trained on 2 tasks: mask language modeling (MLM) (15\% of tokens were masked and next sentence prediction (NSP) (Given the first sentence, BERT was trained to predict whether a selected next sentence was likely or not). BERT is pretrained on Wikipedia and BooksCorpus and needs task-specific fine-tuning. Pretrained BERT models include $BERT_{Base}$ (12 layers, 12 attention heads, and 110 million parameters), as well as $BERT_{Large}$ (24 layers, 16 attention heads, and 340 million parameters).
\begin{table*}[!t]
\centering
\scriptsize
\vspace{-0.25cm}
\caption{Statistics of the datasets used. We used the Stratified 5-Folds cross-validation (CV) strategy for train/test split if original datasets do not have an official train/test split.}
\vspace{-0.35cm}
\label{dsets}
\begin{tabular}{cccccc}
\hline \hline
Task (Classification)                 & Dataset               & Platform      & \# of Samples & \# of Classes & Training Strategy Used \\ \hline \hline
Suicide                               & R-SSD~\cite{cao2019latent}                 & Reddit        & 500 Users     & 5             & Stratified 5-Folds CV                 \\ \hline \hline
\multirow{2}{*}{Stress}               & Dreaddit~\cite{turcan2019dreaddit}
           & Reddit        & 3553 Posts    & 2             & Official Split                 \\
                                      & SAD~\cite{mauriello2021sad}                      & SMS-like      & 6850 SMS      & 2             & Official Split                 \\ \hline \hline
\multirow{4}{*}{Health Mention}       & PHM~\cite{karisani2018did}
             & Twitter       & 4635 Posts    & 4             & Stratified 5-Folds CV                 \\
                                      & PHM~\cite{karisani2018did}               & Twitter       & 4635 Posts    & 2             & Stratified 5-Folds CV                 \\
                                      & HMC2019~\cite{biddle2020leveraging}                 & Twitter       & 15393 Posts   & 3             & Stratified 5-Folds CV                 \\
                                      & RHMD~\cite{naseemHMC}                  & Reddit        & 3553  Posts   & 4             & Stratified 5-Folds CV                 \\ \hline \hline
\multirow{2}{*}{Vaccine Sentiment}    & VS1~\cite{dunn2020limited}
                 & Twitter       & 9261 Posts    & 3             & Stratified 5-Folds CV                 \\
                                      & VS2~\cite{muller2019crowdbreaks}                     & Twitter       & 18522 Posts   & 3             & Stratified 5-Folds CV                 \\ \hline \hline
\multirow{5}{*}{COVID Related}        & Covid Lies~\cite{hossain2020covidlies}            & Twitter       & 3204 Posts    & 3             & Stratified 5-Folds CV                 \\
                                      & Covid Category~\cite{muller2020covid}        & Twitter       & 4328 Posts    & 2             & Stratified 5-Folds CV                 \\
                                      & COVIDSentiA~\cite{naseem2021covidsenti}           & Twitter       & 30000  Posts  & 3             & Stratified 5-Folds CV                 \\
                                      & COVIDSentiB~\cite{naseem2021covidsenti}           & Twitter       & 30000  Posts  & 3             & Stratified 5-Folds CV                 \\
                                      & COVIDSentiC~\cite{naseem2021covidsenti}           & Twitter       & 30000  Posts  & 3             & Stratified 5-Folds CV                 \\ \hline \hline
\multirow{6}{*}{Depression}           & eRISK T3~\cite{losada2016test}              & Reddit        & 190 Users     & 4             & Stratified 5-Folds CV                 \\
                                      & Depression\_Reddit\_1~\cite{naseemdepression}    & Reddit        & 3553 Posts    & 4             & Stratified 5-Folds CV                 \\
                                      & eRISK19 T1~\cite{losada2016test}              & Reddit        & 2810 Users          &     2          & Official Split                 \\
                                      & Depression\_Reddit\_2~\cite{pirina2018identifying} & Reddit        & 1841 Posts    & 2             & Stratified 5-Folds CV                 \\
                                      & Depression\_Twitter\_1 & Twitter       & 1793 Posts    & 3             & Stratified 5-Folds CV                 \\
                                      & Depression\_Twitter\_2 & Twitter       & 10314 Posts   & 2             & Stratified 5-Folds CV                 \\ \hline \hline
\multirow{6}{*}{Other Health related} & PubHealth~\cite{pubhealth}             & News Websites & 12251 Posts   & 4             & Official Split                 \\
                                      & Abortion~\cite{mohammadccsemeval}              & Twitter       & 933 Posts     & 3             & Official Split                 \\
                                      & Amazon Health~\cite{he2016ups}         & Amazon        & 2003 Posts    & 4             & Official Split                  \\
                                      & SMM4H T1~\cite{weissenbacher-etal-2018-overview}              & Twitter       & 14954 Posts   & 2             & Official Split                  \\
                                      & SMM4H T2~\cite{weissenbacher-etal-2018-overview}              & Twitter       & 13498 Posts   & 3             & Official Split                 \\
                                      & HRT~\cite{paul2012model}                   & Twitter       & 2754 Posts    & 4             & Stratified 5-Folds CV                 \\ \hline \hline
\end{tabular}
\vspace{-0.15cm}
\end{table*}

\subsection{Pretraining of PHS-BERT}
We followed the standard pretraining protocols of BERT and initialized PHS-BERT with weights from BERT during the training phase instead of training from scratch and used the uncased version of the BERT model.

PHS-BERT is the first domain-specific LM for tasks related to PHS and is trained on a corpus of health-related tweets that were crawled via the Twitter API. Focusing on the tasks related to PHS, keywords used to collect pretraining corpus are set to disease, symptom, vaccine, and mental health-related words in English. Pre-processing methods similar to those used in previous works~\cite{muller2020covid,nguyen2020bertweet} were employed prior to training. Retweet tags were deleted from the raw corpus, and URLs and usernames were replaced with HTTP-URL and @USER, respectively. Additionally, the Python emoji\footnote{https://pypi.org/project/emoji/} library was used to replace all emoticons with their associated meanings. The HuggingFace\footnote{https://huggingface.co/}, an open-source python library, was used to segment tweets. Each sequence of BERT LM inputs is converted to 50,265 vocabulary tokens. Twitter posts are restricted to 200 characters, and during the training and evaluation phase, we used a batch size of 8. Distributed training was performed on a TPU v3-8.

\subsection{Fine-tuning for downstream tasks}

We applied the pretrained PHS-BERT in the binary and multi-class classification of
different PHS tasks such as stress, suicide, depression, anorexia, health mention classification, vaccine, and covid related misinformation and sentiment analysis. We fine-tuned the PLMs in downstream tasks. Specifically, 
we used the \texttt{ktrain} library  \cite{maiya2020ktrain} to fine-tune each model independently for each dataset. We used the embedding of the special token \texttt{[CLS]} of the last hidden layer as the final feature of the input text. We adopted the multilayer perceptron (MLP) with
the hyperbolic tangent activation function and used Adam optimizer~\cite{kingma2014adam}. The models are trained with a one cycle policy~\cite{smith2017cyclical} at a maximum learning rate of 2e-05 with momentum cycled between 0.85 and 0.95.

\section{Experimental Analysis}\label{results}
\subsection{Tasks and Datasets}

We evaluated and benchmarked the performance of PHS-BERT on 7 different PHS classification tasks (e.g., stress, suicidal ideation, depression, health mention, vaccine,  covid related sentiment analysis, and other health-related tasks) collected from popular social platforms (e.g., Reddit and Twitter). We used 25 datasets (see Table~\ref{dsets}) crawled from social media platforms
(e.g., Reddit and Twitter). We relied on the datasets that are widely used in the community and described each of these tasks and datasets. Below we briefly discussed each task and dataset used in our study (appendix~\ref{appendixxx} for details).

\begin{enumerate}   [leftmargin=*]
    
      \item \textbf{Suicide:} The widespread use of social media for expressing personal thoughts and emotions makes it a valuable resource for assessing suicide risk on social media. We used the following dataset to evaluate the performance of our model.   We used R-SSD~\cite{cao2019latent}  dataset to evaluate the performance of our model on suicide risk detection.
      
      \item \textbf{Stress}: It is desirable to detect stress early in order to address the growing problem of stress. To evaluate stress detection using social media, we evaluated PHS-BERT on the Dreaddit~\cite{turcan2019dreaddit} and SAD~\cite{mauriello2021sad}  datasets.
    
    
     \item \textbf{Health mention:} In social media platforms, people often use disease or symptom terms in ways other than to describe their health. In data-driven PHS, the health mention classification task aims to identify posts where users discuss health conditions rather than using disease and symptom terms for other reasons. We used PHM~\cite{karisani2018did}, HMC2019~\cite{biddle2020leveraging} and RHMD\footnote{https://github.com/usmaann/RHMD-Health-Mention-Dataset}  health mention-related datasets.
     
      

\item \textbf{Vaccine sentiment:} Vaccines are a critical component of public health. On the other hand, vaccine hesitancy and refusal can result in clusters of low vaccination coverage, diminishing the effectiveness of vaccination programs. Identifying vaccine-related concerns on social media makes it possible to determine emerging risks to vaccine acceptance. We used VS1~\cite{dunn2020limited} and VS2~\cite{muller2019crowdbreaks}  vaccine-related Twitter datasets to show the effectiveness of our model.

    \item \textbf{COVID related}: Due to the ongoing pandemic, there is a higher need for tools to identify COVID-19-related misinformation and sentiment on social media. Misinformation can have a negative impact on public opinion and endanger the lives of millions of people if precautions are not taken. We used COVID Lies~\cite{hossain2020covidlies}, Covid category~\cite{muller2020covid}, and COVIDSenti~\cite{naseem2021covidsenti}\footnote{we used 3 subsets (COVIDSentiA, COVIDSentiB and COVIDSentiC)} datasets to test our model.
    
     \item \textbf{Depression}: User-generated text on social media has been actively explored for its feasibility in the early identification of depression. We used following eRisk T3~\cite{losada2016test}, eRisk T1~\cite{losada2016test},  Depression\_Reddit\_1~\cite{naseemdepression}\footnote{https://github.com/usmaann/Depression\_Severity\_Dataset}, Depression\_Reddit\_2~\cite{pirina2018identifying}, Depression\_Twitter\_1\footnote{https://github.com/AshwanthRamji/Depression-Sentiment-Analysis-with-Twitter-Data}, and Depression\_Twitter\_2\footnote{https://github.com/viritaromero/Detecting-Depression-in-Tweets} depression-related datasets in our experiments.
    
     \item \textbf{Other health related tasks:} We also evaluated the performance of our PHS-BERT on other health-related 6 datasets. We used  PUBHEALTH~\cite{pubhealth}, Abortion~\cite{mohammadccsemeval}\footnote{The SemEval 2016 stance detection task has 5 target domains. We used the legalization of abortion.}, Amazon Health dataset~\cite{he2016ups},  SMM4H T1~\cite{weissenbacher-etal-2018-overview}, SMM4H T2~\cite{weissenbacher-etal-2018-overview} and HRT~\cite{paul2012model}.
    \end{enumerate}

\subsection{Evaluation Metric}
To evaluate the performance, we used  F1-score and the relative improvement in marginal performance \((\Delta MP)\) used in a previous similar study~\cite{muller2020covid}.



\subsection{Baselines}
We evaluated the performance of PHS-BERT with various SOTA existing PLMs in different domains. We compared the performance with BERT~\cite{devlin-etal-2019-bert}, ALBERT~\cite{lan2019albert}, and DistilBERT~\cite{sanh2019distilbert} pretrained with general corpus, BioBERT~\cite{lee2019biobert} pretrained in the biomedical domain, CT-BERT~\cite{muller2020covid} and BERTweet~\cite{nguyen2020bertweet} pretrained on covid related tweets  and MentalBERT~\cite{ji2021mentalbert} pretrained on    corpus from Reddit from mental
health-related subreddits. 

\begin{table*}[!t]
\centering
\scriptsize
\vspace{-0.35cm}
\caption{Comparison of PHS-BERT (Ours) v/s SOTA PLMs. Best results (F1-score) are represented in bold, whereas second-best results are underlined. $\Delta$\(MP_{BERT}\) and $\Delta$\(MP_{SB}\) represent the marginal increase in performance compared to the BERT and the second-best PLM (under-lined).}
\vspace{-0.35cm}
\label{results}
\hspace*{-1.8cm}
\begin{tabular}{ccccccccccc}
\hline \hline
\multicolumn{11}{c}{\textbf{Suicide Ideation Task}}                                                                               \\ \hline \hline
Dataset               & BERT  & ALBERT & distilBERT & CT-BERT & BioBERT & BERTweet & MentalBERT & Ours & $\Delta$$MP_{BERT}$ & $\Delta$$MP_{SB}$    \\ \hline \hline
R-SSD~\cite{cao2019latent}                 & 25.72 & 23.07  & \underline{26.96}      & 18.67   & 23.51   & 24.82    & 17.35      & \textbf{30.28}   & 18.45$\uparrow$        & 12.79$\uparrow$ \\ \hline \hline
\multicolumn{11}{c}{\textbf{Stress   Detection Task}}                                                                             \\ \hline \hline
Dreaddit~\cite{turcan2019dreaddit}              & 78.55 & 79.43  & 78.22      & \underline{81.46}   & 78.34   & 80.03    & 80.89      & \textbf{82.89}          & 5.60$\uparrow$ & 1.78$\uparrow$      \\
SAD~\cite{mauriello2021sad}                    & 92.66 &91.11   &91.47       &91.11    &93.92    &\underline{94.17}    &93.23       & \textbf{94.75}          & 2.28$\uparrow$ &    0.62$\uparrow$   \\ \hline \hline
\textbf{Average}               &85.61  &85.27   &84.85       &86.29    &86.13    &87.10     &  87.06     & \textbf{88.82}         & 3.80$\uparrow$  &     2.00$\uparrow$  \\ \hline \hline
\multicolumn{11}{c}{\textbf{Health Mention Task}}                                                                               \\ \hline \hline
PHM   (Multi-class)~\cite{karisani2018did}   & 86.21 & 80.05  & 85.06      & 82.02   & 82.22   & 85.59    & \underline{87.76}      & \textbf{89.38}         & 3.72$\uparrow$   & 1.87$\uparrow$  \\
PHM (Binary)~\cite{karisani2018did}          &  91.89     &   90.53     &    90.64        & 92.17        &  89.62       & 92.12         &     \underline{92.29}       &   \textbf{93.27}          & 1.52$\uparrow$     &  1.07$\uparrow$      \\
HMC2019~\cite{biddle2020leveraging}             & 88.99 & 87.22  & 88.01      & \underline{90.82}   & 86.27   & 90.65    & 90.17      & \textbf{91.71}        & 3.09$\uparrow$    & 0.99$\uparrow$  \\
RHMD                  &  74.20     &   69.02     &   73.22         &    72.87     &    72.25     &    74.66      &  \underline{75.28}          &  \textbf{77.16}            & 5.48$\uparrow$    &  2.53  $\uparrow$   \\ \hline \hline
\textbf{Average}               &  85.07     &     81.71   &   84.23         &    84.47     & 82.59        &   85.76       &  86.38          &      \textbf{87.38 }       & 3.34$\uparrow$    &  1.76 $\uparrow$    \\ \hline \hline
\multicolumn{11}{c}{\textbf{Depression   Detection Task}}                                                                          \\ \hline \hline
eRisk T3~\cite{losada2016test}               &  64.56     &64.78        &67.33            &63.17         & 64.86        & 63.56         &\underline{67.75}            &     \textbf{68.98  }  & 6.95$\uparrow$         & 1.84$\uparrow$       \\
Depression\_Reddit\_1 & 22.39      &21.09        &21.95            &\underline{24.21}         &24.00         &20.84          &21.95            &  \textbf{28.75}            & 29.73$\uparrow$    & 19.56$\uparrow$        \\
eRisk T1~\cite{losada2016test}               & 93.72      & 93.79        & 93.34           &86.74         &91.73         &91.92          &\underline{94.30}            &\textbf{94.52}        & 0.86$\uparrow$          &0.24$\uparrow$        \\
Depression\_Reddit\_2~\cite{pirina2018identifying} & 91.33 & 90.72  & 91.01      & 68.16   & 90.53    & 91.75    & \underline{92.70}      & \textbf{93.36}        & 2.25$\uparrow$    & 0.72$\uparrow$      \\
Depression\_Twitter\_1         & 64.17 & 51.70   & 66.71      & 57.11   & 64.12   & 64.24    & \underline{72.95}      & \textbf{76.18 }    & 19.01$\uparrow$       & 4.49$\uparrow$  \\
Depression\_Twitter\_2          & 96.99 & 96.79  & 96.70       & 96.96   & 96.59   & 96.87    & \underline{97.09}      & \textbf{97.12}    & 0.14$\uparrow$        & 0.03$\uparrow$  \\ \hline \hline
\textbf{Average}               & 72.19      & 69.81       & 72.84           &66.06            & 71.97         & 71.53          &74.46              &  \textbf{76.49}         & 6.03$\uparrow$        &    2.76$\uparrow$   \\ \hline \hline
\multicolumn{11}{c}{\textbf{Vaccine   Sentiment  Task}}                                                                            \\ \hline \hline
VS1~\cite{dunn2020limited}
                & 74.14 & 70.00  & 73.95      & \underline{79.92}   & 73.30   & 76.81    & 71.56      & \textbf{79.96 }   & 7.96$\uparrow$        & 0.05$\uparrow$  \\
VS2~\cite{muller2019crowdbreaks}                      & 76.60 & 74.82  & 75.91      & \underline{81.73}   & 76.77   & 79.10    & 77.65      & \textbf{82.24}         & 7.46$\uparrow$   & 0.63$\uparrow$  \\ \hline \hline
\textbf{Average}               & 75.37 & 72.41  & 74.93      & 80.84   & 75.04   & 77.96    & 74.61      & \textbf{81.10 }     & 7.70$\uparrow$      & 0.34$\uparrow$  \\ \hline \hline
\multicolumn{11}{c}{\textbf{COVID Related  Task}}                                                                                  \\ \hline \hline
Covid Lies~\cite{hossain2020covidlies}

           & 92.96 & 91.53  & 92.14      & 92.24   & 93.79   & 91.07    &\underline{ 94.60}      & \textbf{95.35 }      & 2.60$\uparrow$     & 0.80$\uparrow$  \\
COVID   Category~\cite{muller2020covid}   & 93.98 & 93.94  & 94.35      & \underline{95.29}   & 93.72   & 93.45    & 94.97      & \textbf{95.83}         & 1.99$\uparrow$   & 0.57$\uparrow$  \\
COVIDSentiA~\cite{naseem2021covidsenti}            & 90.90 & 90.81  & 90.90      & 78.96   & 90.41   & 66.30    & \underline{91.55}      & \textbf{93.97}      & 3.41$\uparrow$      & 2.67$\uparrow$  \\
COVIDSentiB~\cite{naseem2021covidsenti}            & 91.31 & 89.88  & 91.06      & 86.85   & 91.02   & 89.46    & \underline{92.06}      & \textbf{93.44 }       & 2.36$\uparrow$    & 1.52$\uparrow$  \\
COVIDSentiC~\cite{naseem2021covidsenti}            & 91.24 & 83.72  & 90.77      & 84.83   & 90.55   & 61.78    & \underline{91.66}      & \textbf{93.11 }        & 2.03$\uparrow$   & 1.60$\uparrow$  \\ \hline \hline
\textbf{Average}               & 92.08 & 89.98  & 91.84      & 87.63   & 91.90   & 80.41    & 92.97      & \textbf{94.34}        & 2.48$\uparrow$    & 1.49$\uparrow$  \\ \hline \hline
\multicolumn{11}{c}{\textbf{Other Health   Related Task}}                                                                          \\ \hline \hline
PubHealth~\cite{pubhealth}             & 60.30 & 61.43  & 60.77      & \underline{63.97}   & 58.85   & 60.57    & 57.30      & \textbf{64.77}          & 7.54$\uparrow$  & 1.27$\uparrow$  \\
Abortion~\cite{mohammadccsemeval}              & 58.79 & 58.59  & 68.09      & \underline{70.39}   & 62.53   & 62.82    & 63.03      & \textbf{72.31}       & 23.40$\uparrow$     & 2.77$\uparrow$  \\
Amazon Health~\cite{he2016ups}         & 63.45 & 63.18  & 62.30      & 54.84   & 60.27   & 65.50    & \underline{65.57}      & \textbf{68.09 }     & 7.43$\uparrow$      & 3.90$\uparrow$  \\
SMM4H T1~\cite{weissenbacher-etal-2018-overview}              & 33.33 & 33.86  & 35.80     & 45.50   & 39.45   & \underline{45.87}    & 39.81      & \textbf{46.49}      & 40.71$\uparrow$      & 1.38$\uparrow$      \\
SMM4H T2~\cite{weissenbacher-etal-2018-overview}              & 75.54 & 72.76  & 75.12      & 79.19   & 73.43   & \underline{80.20}    & 77.54      & \textbf{80.34}        & 6.44$\uparrow$    &    0.18$\uparrow$   \\
HRT~\cite{paul2012model}                   & 78.67 & 76.97  & 78.35      & \underline{80.90}   & 76.13   & 80.48    & 80.46      & \textbf{81.12}         & 3.15$\uparrow$   &  0.28$\uparrow$      \\ \hline  \hline
\textbf{Average}               & 61.68 & 61.13  & 63.41      & 65.80   & 61.78   & 65.91    & 63.95      & \textbf{68.85}        & 11.82$\uparrow$    &  4.71$\uparrow$     \\ \hline \hline
\end{tabular}
\vspace{-0.35cm}
\end{table*}

\subsection{Results}

Table~\ref{results} summarizes the results of the presented PHS-BERT in comparison to the baselines. We observe that the performance of PHS-BERT is higher than SOTA PLMs on all tested tasks and datasets. Below we discuss the performance comparison of PHS-BERT with BERT and the results of the second-best PLM.




\noindent \textbf{Suicide Ideation Task:} We observed that the marginal increases in performance of PHS-BERT is 18.45\% when compared to BERT and 12.79\% when compared to second best results.


\noindent \textbf{Stress Detection Task:} We showed that PHS-BERT achieved higher performance than the best baseline on both datasets. The average marginal increase in performance of PHS-BERT is 3.80\% compared to BERT and 2\% when compared to second-best results.


\noindent \textbf{Health Mention Task:} PHS-BERT outperformed all the baselines on all health mention classification datasets. The average marginal increase in performance of PHS-BERT is 3.34\% compared to BERT and 1.76\% when compared to second-best results.


\noindent \textbf{Depression Detection Task:} We demonstrated that PHS-BERT outperformed all the baselines on all 6 depression datasets to identify depression on social media. We observed that the average marginal increase in performance of PHS-BERT is 6.03\% compared to BERT and 2.76\% when compared to second-best results.


\noindent \textbf{Vaccine Sentiment Task:} For the vaccine sentiment task, PHS-BERT achieved higher performance compared to all baselines on both datasets.  Results showed that the average marginal increase in performance of PHS-BERT is 7.70\% than BERT and 0.34\% compared to second-best results.

\noindent \textbf{COVID Related Task:} PHS-BERT outperformed all baselines on all 5 datasets for COVID-related tasks. On average, the marginal increase in performance is 11.82\% compared to BERT and 4.471\% compared to the second-best results.


\noindent \textbf{Other Health Related Task:}  We showed that PHS-BERT outperformed all the baselines on all 6 datasets to identify other health-related tasks on social media. We observed that the average marginal increase in performance of PHS-BERT is 11.82\% compared to BERT and 4.71\% when compared to second-best results.


\subsection{Discussion}

We demonstrated the effectiveness of our domain-specific PLM on a downstream classification task related to PHS. Compared to previous SOTA PLMs, PHS-BERT
improved the performance on all datasets (7 tasks). Our experimental results showed that BERT, a PLM trained in the general domain, gets competitive results on downstream classification tasks. However, for domain-specific tasks, general domain PLMs (BERT, ALBERT, distilBERT) might need more training on relevant corpora to achieve better performance on the domain-specific downstream classification task. Further, we observed that using a domain-specific PLM trained on biomedical corpora (BioBERT) is less effective than pretraining on the target domain. We also observed that using CT-BERT, BERTweet, and MentalBERT, which are trained on social media-based text, performs better compared to PLMs trained in the general and biomedical domain. These results also demonstrated the effectiveness of training in a target domain. In particular, CT-BERT has the second-best performance on 9 datasets, and MentalBERT has the second-best performance on 13 datasets. The results of domain-specific PLMs demonstrated that continued pretraining in the relevant domain improves performance on downstream tasks.

\section{Conclusion}\label{con}

We present PHS-BERT, a domain-specific PLM trained on health-related social media data. Our results demonstrate that using domain-specific corpora to train general domain LMs improves performance on PHS tasks. On all 25 datasets related to 7 different PHS tasks, PHS-BERT outperforms previous state-of-the-art PLMs. We expect that the PHS-BERT PLM will benefit the development of new applications based on PHS NLP tasks.
\noindent\textbf{Ethics and Societal Impact} 

\noindent \textbf{Ethics:} No additional ethics approval was sought for the analysis of data in this study because data were drawn from already published studies. 

\noindent\textbf{Societal Impact:} We train and release a PLM to accelerate the automatic identification of tasks related to PHS on social media. Our work aims to develop a new computational method for screening users in need of early intervention and is not intended to use in clinical settings or as a diagnostic tool.

\noindent\textbf{Reproducibility:} 
For reproducibility and future works, PHS-BERT is publicly released and is available at \url{https://huggingface.co/publichealthsurveillance/PHS-BERT}. 


\bibliography{anthology,custom}
\bibliographystyle{acl_natbib}
\clearpage
\appendix

\section{Dataset description}
\label{appendixxx}

\begin{enumerate}   [leftmargin=*]
    \item \textbf{Depression}: We used 6 depression-related datasets in our experiments. 
    \begin{itemize} [leftmargin=*]
        \item \textbf{eRisk T3}: We used eRISK, a publicly available dataset, released by~\cite{losada2016test} and labeled across 4 depression severity levels using Beck’s Depression Inventory~\cite{beck1961inventory} criteria to detect the existence of depression and identify its severity level in social media posts. eRISK was later used in the CLEF's eRISK challenge Task 3\footnote{https://erisk.irlab.org/2021/index.html}  on early identification of depression in social media. Since in each years' challenge author released a small number of user's data (ranging from 70-90 users data), we combined and used the data of the last 3 years, which is equivalent to 190 Reddit users, labeled across 4 depression severity levels.
        \item \textbf{Depression\_Reddit\_1}: We used new Reddit depression data released by \citet{naseemdepression}. This dataset consists of 3,553 Reddit posts to identify the depression severity on social media. Annotators manually labeled data into 4 depression severity levels i.,e., (i) minimal depression; (ii) mild depression, (iii) moderate depression; and (iv) severe depression using  Depressive Disorder Annotation scheme~\cite{mowery2015towards}.


\item \textbf{eRisk T1}: The third depression data is from eRisk shared task 1~\cite{losada2016test}, which is a public competition for detecting early risk in health-related areas. The eRisk data consists of posts from 2,810 users, with 1,370 expressing depression and 1,440 as a control group without depression.

\item \textbf{Depression\_Reddit\_2}: 
The fourth depression dataset used is released by Pirina and Çöltekin~\cite{pirina2018identifying}. The authors used Reddit to collect additional social data, which they combined with previously collected data to identify depression.

\item \textbf{Depression\_Twitter\_1}: Our fifth depression dataset is a publicly availabl\footnote{https://github.com/AshwanthRamji/Depression-Sentiment-Analysis-with-Twitter-Data}. This data is collected from Twitter and labeled into 3 labels (e.g., Positive, Negative, and Neutral) for depression sentiment analysis.
\item \textbf{Depression\_Twitter\_2}: Our sixth depression dataset is a public dataset\footnote{https://github.com/viritaromero/Detecting-Depression-in-Tweets}, collected from Twitter and labeled into 2 labels (e.g., Positive and Negative) for depression detection.

    \end{itemize}


    
    \item \textbf{Health Mention:} We used 3 health mention-related datasets in our experiments.
    \begin{itemize} [leftmargin=*]
        \item \textbf{PHM}: \citet{karisani2018did}  constructed and released the PHM dataset consisting of 7,192 English tweets across 6 diseases and symptoms. They used the Twitter API to retrieve the data using the colloquial disease names as search keywords. They manually annotated the tweets and categorized them into 4 labels. In addition to 4 labels, similar to \citet{karisani2018did} we also used binary labels for health mention classification. 
        
        \item \textbf{HMC2019}: HMC2019 is presented by \citet{biddle2020leveraging} by extending the PHM dataset to include 19,558  tweets and included labels related to figurative mentions, and included  4 more different disease or symptom terms (10 in total) for health mention classification.
        
        \item \textbf{RHMD}: We also used Reddit health mention dataset (RHMD)\cite{naseemHMC} for HMC task. RHMD consists of 10K+ Reddit posts manually annotated with 4 labels (personal health mention, non-personal health mention, figurative health mention, hyperbolic health mention). In our study, we used 3 label versions of data released by authors where they merged figurative health mention and hyperbolic health mention into 1 class.
        
    \end{itemize} 
    
     \item \textbf{Suicide:} We used the following dataset to evaluate the performance of our model on suicide risk detection.

    \begin{itemize} [leftmargin=*]
    \item \textbf{R-SSD}: For suicide ideation, we used a dataset released by \citet{cao2019latent}, which contains 500 individuals' Reddit postings categorized into 5 increasing suicide risk classes from 9 mental health and suicide-related subreddits. 
    \end{itemize}
    \item \textbf{Stress}: To evaluate stress detection using social media, we evaluated PHS-BERT on the following datasets.
    \begin{itemize}
    \item \textbf{Dreaddit}: For stress detection, we used Dreaddit~\cite{turcan2019dreaddit} collected from 5 different Reddit forums. Dreaddit consists of 3,553 posts and focuses on three major stressful topics: interpersonal conflict, mental illness, and financial need. Posts in Dreaddit are collected from 10 subreddits, including some mental health domains such as anxiety and PTSD. 
    \item \textbf{SAD}:  The SAD~\cite{mauriello2021sad} dataset, which contains 6,850 SMS-like sentences, is used to recognize everyday stressors. The SAD dataset is derived from stress management articles, chatbot-based conversation systems, crowdsourcing, and web crawling. Some of the more specific stressors are work-related issues like fatigue or physical pain, financial difficulties like debt or anxiety, school-related decisions like final projects or group projects, and interpersonal relationships like friendships and family relationships.

    
    \end{itemize}

    \item \textbf{Vaccine sentiment:} We used two vaccine-related Twitter datasets to show the effectiveness of our model. 
    \begin{itemize} [leftmargin=*]
    \item \textbf{VS1}:  Our first dataset consists of tweets about vaccine dissemination on Twitter from January 12, 2017, to December 3, 2019. \citet{dunn2020limited} crawled and labeled this data. The total tweets count is 9,212, with 6,683 positive, 1,084 negatives, and 1,445 neural tweets.
    \item \textbf{VS2}: The second dataset\footnote {https://github.com/digitalepidemiologylab/crowdbreaks-paper} includes tweets about measles and vaccinations obtained via the Twitter Streaming API between July 2018 and January 2019 and provided by \citet{muller2019crowdbreaks}. The total number of tweets is 18,503, with 8,965 pro-vaccine tweets, 1,976 anti-vaccine tweets, and 7,562 neutral tweets.
    \end{itemize}

    \item \textbf{COVID}: We used 5 covid related datasets to test our model.
    \begin{itemize} [leftmargin=*]
     \item \textbf{COVID Lies}:  \citet{hossain2020covidlies} released COVIDLIES, a dataset (6761 tweets) annotated by experts with known COVID-19 misconceptions and tweets that agree, disagree, or express no stance.
    \item \textbf{Covid category}: Covid category dataset is released by \citet{muller2020covid}. Amazon Turk  annotators were asked to classify a given tweet text as personal narrative or news. Crowdbreaks was used to perform the annotation.
    \item \textbf{COVIDSenti}: We used a newly released large-scale sentiment dataset, COVIDSenti, which contains 90,000 COVID-19-related tweets obtained during the pandemic's early stages, from February to March 2020. The tweets are labeled into positive, negative, and neutral sentiment classes. In our experiments, we used 3 subsets (COVIDSentiA, COVIDSentiB and COVIDSentiC) released by authors~\cite{naseem2021covidsenti}.
   
    \end{itemize}

    \item \textbf{Other health related tasks:} We used  PUBHEALTH~\cite{pubhealth}, a dataset for automated fact-checking of public health claims that are explainable. PUBHEALTH is labeled with its factuality (true, false, unproven, mixture). (ii) Abortion:   In SemEval 2016 stance
detection task~\cite{mohammadccsemeval},  5 target domains are given: legalization of abortion, atheism, climate change, feminism, and Hillary Clinton.  We used the legalization of abortion in our experiments. (iii) Amazon Health dataset:  The Amazon Health dataset~\cite{he2016ups} contains reviews of Amazon healthcare products and has 4 classes i.e., strongly positive, positive, negative, and strongly negative. (iv)  SMM4H T1: We used Social Media Mining for Health (SMM4H) Shared  Task 1 recognizing whether a tweet is reporting an
adverse drug reaction~\cite{weissenbacher-etal-2018-overview}.  (v) SMM4H T2:  Drug Intake Classification (SMM4H Task 2)~\cite{weissenbacher-etal-2018-overview} where participants were given tweets manually categorized as definite intake, possible intake, or no intake. (vi) HRT: Health related tweets (HRT)~\cite{paul2012model} were collected using Twitter and manually annotated  using Mechanical Turk as related or unrelated to health. Health-related tweets were further labeled as sick (the text implied that the user was suffering from an acute illness, such as a cold or the flu) or health (the text made general comments about the user's or the other's health, such as chronic health conditions, lifestyle, or diet) and unrelated tweets were further labeled as unrelated (texts that were not about a specific person's health, such as news and updates about the swine flu or advertisements for diet pills) and non-English.
    \end{enumerate}

\end{document}